\documentclass[10pt,twocolumn,letterpaper]{article}

\usepackage{iccv}
\usepackage{times}
\usepackage{epsfig}
\usepackage{graphicx}
\usepackage{amsmath,amssymb}
\usepackage{color}
\usepackage{latex_pkgs/maths}
\usepackage{latex_pkgs/msformatting}
\usepackage{algorithm2e}
\usepackage{listings}
\usepackage{cuted} % for strip
\SetAlFnt{\small}

\usepackage[pagebackref=true,breaklinks=true,letterpaper=true,colorlinks,bookmarks=false]{hyperref}

\iccvfinalcopy % *** Uncomment this line for the final submission

\def\iccvPaperID{1431} % *** Enter the ICCV Paper ID here

\ificcvfinal\fi

\begin{document}
%%%%%%%%% TITLE
\title{Online Real-time Multiple Spatiotemporal Action Localisation and Prediction}
%\and
\author{
Gurkirt Singh${}^1$
\qquad Suman Saha${}^1$
\qquad Michael Sapienza${}^2$\thanks{
  M. Sapienza performed this research at the University of Oxford, and is currently with the Think Tank Team, Samsung Research America, CA.} \qquad Philip Torr${}^2$ \qquad Fabio Cuzzolin${}^1$ \\ 
${}^1$Oxford Brookes University \qquad ${}^2$University of Oxford\\ 
{\tt\small \{gurkirt.singh-2015,suman.saha-2014,fabio.cuzzolin\}@brookes.ac.uk}
\\
{\tt\small m.sapienza@samsung.com, philip.torr@eng.ox.ac.uk}
%\{michael.sapienza,philip.torr\}@eng.ox.ac.uk} 
}

\maketitle
\thispagestyle{empty} % removed comment to remove the page number from the first page
\begin{strip}
  \vskip -1cm
  \centering
  \includegraphics[width=0.85\textwidth]{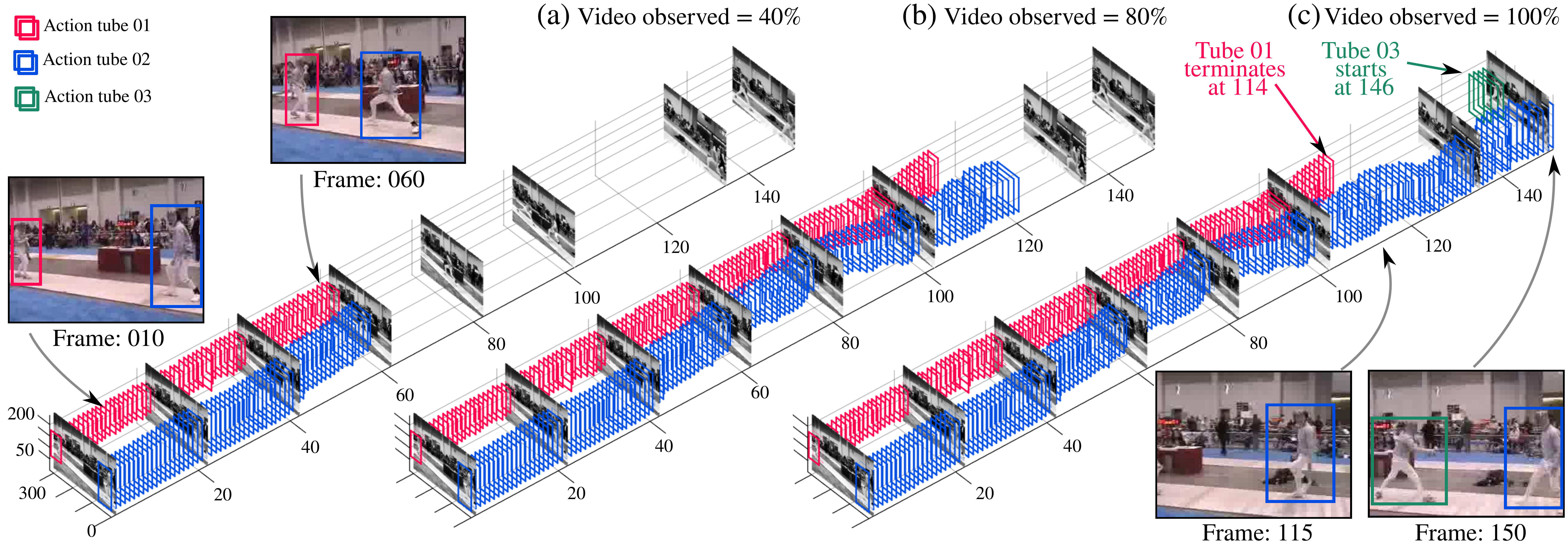}
  \vskip 0.3cm
  \begin{minipage}[adjusting]{0.92\textwidth}
  {\fontsize{3.2mm}{1mm}\selectfont
    Figure 1: Online spatio-temporal action localisation 
    %and early action label prediction 
    in a test `fencing' video from UCF-101-24~\cite{soomro-2012}.
    \textbf{(a)} to \textbf{(c)}:  A 3D volumetric view of the video showing detection boxes and selected frames.
    At any given time, a certain portion (\%) of the entire video is observed by the system, 
    and the detection boxes are linked up to incrementally build space-time action tubes.} %in real-time.}
   Note that the proposed method is able to detect multiple co-occurring action instances (3 tubes shown here). 
   %(3 action instances are shown in different colours).
   %Note also that one of the fencers moves out of the image boundaries between frames 114 and 145,
   %to which our model responds by trimming action tube 01 at frame 114, and initiating a new tube (03) at frame 146.}
  \end{minipage}
  \label{fig:introductionTeaser}
\end{strip}
\setcounter{figure}{1}
%%%%%%%%% ABSTRACT
\begin{abstract}  
We present a deep-learning framework for real-time multiple spatio-temporal (S/T) action localisation and classification.
Current state-of-the-art approaches work %in an offline mode
offline, and are too slow to be useful in real-world settings.
To overcome their limitations we introduce two major developments. 
Firstly, we adopt real-time SSD (Single Shot MultiBox Detector) 
CNNs
%convolutional neural networks
to regress and classify
detection boxes in each video frame potentially containing an action of interest.
Secondly, we design an original and efficient online %`action tube' 
%generation method 
algorithm to incrementally construct and label `action tubes' from the SSD frame level detections. % to construct action tubes incrementally and labelling action tubes at the same time.
As a result, our system is not only capable of performing S/T detection in real time, but can also perform early action prediction in an online fashion. 
We achieve new state-of-the-art results in both S/T action localisation and early action prediction on the challenging UCF101-24 and J-HMDB-21 benchmarks,
even when compared to the top offline competitors. 
To the best of our knowledge, 
ours is the first real-time (up to 40fps) system
able to perform online S/T action localisation on the untrimmed videos of UCF101-24.
%Code is avaiable online\url{https://github.com/gurkirt/realtime-action-detection}.
%\GUR{and able to predict the label of the untrimmed videos.}
\iffalse
%CVPR ABSTRACT
We present a method for multiple spatio-temporal (S/T) action localisation, 
classification and early prediction based on a single deep learning framework, 
able to perform in an online fashion and in real time.
Our online action detection pipeline comprises of three main steps. 
In step one, two end-to-end trainable SSD (Single Shot MultiBox Detector) 
convolutional neural networks are employed to regress and classify
detection boxes in each video frame potentially containing an action of interest, 
one from optical flow and one from RBG images. In step two, appearance and motion 
cues are combined by merging the detection boxes and classification
scores generated by the two networks. In step three, sequences of detection boxes most 
likely to be associated with a single action instance, called ‘action tubes’, 
are constructed incrementally and efficiently in an online fashion, allowing
the system to perform early action class prediction and spatial localisation in real time. 
Empirical results on the challenging UCF101 and J-HMDB-21 datasets demonstrate
new state-of-the-art results in S/T action localisation
on UCF101, even when compared to offline competitors,
and superior results on early action label prediction across
the board. To the best of our knowledge, 
ours is the first real-time framework (up to 40fps) 
able to perform online spatio-temporal action localisation.
\fi

%\vskip -1cm
\end{abstract}
%%%%%%%%% BODY TEXT
%\iffalse
% \begin{figure}[t]
%   %\centering
% \includegraphics[width=0.99\textwidth]{figures/cvpr2017IntroductionTeaser.pdf}
% \caption{
%  Action prediction and online action localisation in a `biking' test video taken from UCF-101~\cite{J-HMDB-Jhuang-2013} dataset.  
%   \textbf{(a)} to \textbf{(c)}:  A 3D volumetric view of the video with selected frames.
%   At any given time, a certain portion (\%) of the entire video is observed by the system, 
%   and the detection boxes are linked up to incrementally build an online space-time action tube.  

  % A video sequence taken from the LIRIS-HARL dataset plotted in space-and time.
%     \textbf{(a)} A top down view of the video plotted with the detected action tubes 
%     of class `handshaking' in green, and `person leaves baggage unattended' in red. 
%     Each action is located to be within a space-time tube.
%     \textbf{(b)} A side view of the same space-time detections. 
%     Note that no action is detected at the beginning of the video when 
%     there is human motion present in the video.
%     \textbf{(c)} The detection and instance segmentation result of 
%     two actions occurring simultaneously in a single frame(make chnages)
%     }
%   \label{fig:introductionTeaser}
% \end{figure}
%\fi
%%%%%%%%% BODY TEXT
\section{Introduction} \label{sec:intro}
% ----------------------- online action detection ---------------------------------------------------------------------
Spatio-temporal human action localisation~\cite{Weinzaepfel-2015,Saha2016,peng2016eccv} in videos is a challenging problem
that is made even harder if detection is to be performed in an online setting and at real-time speed.
% [SUMAN] I have removed a line break at this point - do we need a para here?
Despite the performance of state-of-the-art S/T action detection systems~\cite{Saha2016,peng2016eccv} being far from real time,
current systems also assume that the entire video (taken as a 3D block of pixels) is available ahead of time in order to detect action instances.
Here, an action instance is made up of a sequence of detection boxes linked in time to form an `action tube'~\cite{Georgia-2015a,Weinzaepfel-2015}.
For such a detector to be applicable to real-world scenarios such as video surveillance and human-robot interaction,
video frames need to be processed in real time.
Moreover, the action detection system needs to construct action tubes in an incremental and online fashion, as each new frame is captured.
%human robot interaction (e.g. surgical robotics) and other applications requiring a prompt response,

\iffalse
% ----------------- brief statement of paper contributions -------------------------------------------------------
%%%%%%REMOVED BY GUR
%\emph{Contribution}.
In this work, we design and demonstrate a deep learning framework able to perform spatial and temporal action localisation and classification in an online fashion (see Fig.~1) and in real-time, while further
improving the detection accuracy of the latest state-of-the-art offline action detectors~\cite{Saha2016,peng2016eccv} 
on the most complex available benchmark.
%%%\iffalse
As it works by constructing action tubes in a fully incremental way, as illustrated in Fig.~1,
the framework can also perform early prediction of action tube labels.
\fi
% ------------------------------  latest related developments ----------------------------------------------------
%\emph{Current state of the art}.
With the rise of Convolutional Neural Networks (CNNs),
impressive progress has been made in image classification \cite{krizhevsky2012} and object detection \cite{girshick-2014},
motivating researchers to apply CNNs to action classification and localisation.
Although the resulting CNN-based state-of-the-art S/T action detectors~\cite{Saha2016,Georgia-2015a,Weinzaepfel-2015,peng2016eccv} have achieved
remarkable results, these methods are computationally expensive and their detection accuracy is still below what is needed 
for real-world deployment.
%especially on realistic detection thresholds of $\delta = 0.5$ or more.
Most of these approaches~\cite{Georgia-2015a,Weinzaepfel-2015} are based on unsupervised 
region proposal algorithms~\cite{uijlings-2013,zitnick2014edge} 
and on an expensive multi-stage training strategy mutuated from object detection~\cite{girshick-2014}. 
For example, Gkioxari~\etal~\cite{Georgia-2015a} and Weinzaepfel~\etal~\cite{Weinzaepfel-2015}
both separately train a pair of (motion and appearance) CNNs and a battery of one-vs-rest Support Vector Machines (SVMs). 
This limits detection accuracy as each module is trained independently, leading to sub-optimal solutions.
\iffalse
Furthermore, multi-stage algorithms~\cite{Georgia-2015a,Weinzaepfel-2015}
exhibit high computational and storage requirements, which prevent them from achieving real-time capabilities.
\fi

The most recent efforts by Saha \etal ~\cite{Saha2016} and Peng \etal \cite{peng2016eccv} use a supervised region proposal generation 
approach~\cite{ren2015faster}, and eliminate the need for multi-stage training~\cite{girshick-2014} by using 
a single end-to-end trainable CNN for action classification and bounding box regression. 
%Their results show that supervised region proposal generation is crucial to improve detection accuracy.
%as they can help filter large numbers of region proposals to select only those which are highly likely to contain an action.
{Although \cite{Saha2016,peng2016eccv} exhibit the best spatio-temporal action localisation accuracies to date, test time
detection involves the use of computationally expensive optical flow ~\cite{Brox-2004}, and remains a two-step region proposal network (RPN)~\cite{ren2015faster} and RCNN~\cite{ren2015faster} process, limiting real-time deployment.
Also, \cite{Saha2016,peng2016eccv} both employ offline tube generation methods which process the entire video in two passes:
one to link detection boxes into tubes which stretch from start to end of the video,
and one to temporally trim and label the video-long constructed tubes.}
%Their use of  non-real-time optical flow ~\cite{Brox-2004} is an issue from a real-time perspective.
%Finally, these methods' mAP at high detection overlap values remains disappointingly low, and incompatible with many real world applications.}

% ---------------------------------- how do we fix those issues with competitors -------------------------------
%\emph{Features of our framework}.

\begin{figure*}[t]
  \centering
  \includegraphics[scale=0.32]{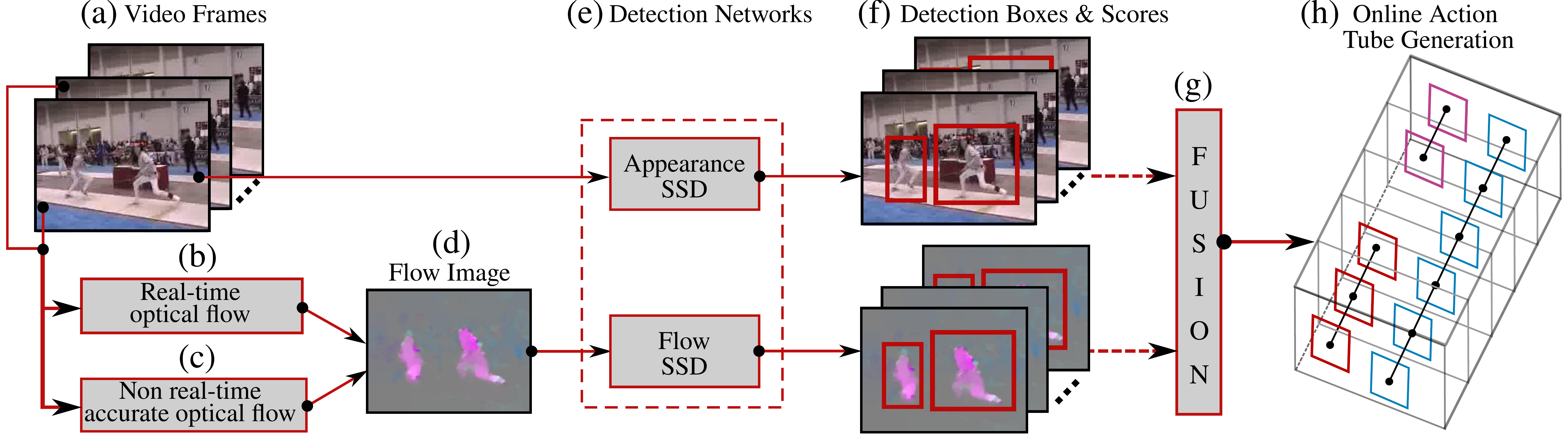}
  \vskip 2mm
  \caption{
    {\small
      \textit{
	    At test time, the input to the framework is a sequence of RGB video frames \textbf{(a)}.
	    A real-time optical flow (OF) algorithm \textbf{(b)} \cite{kroeger2016fast} takes the consecutive RGB frames as input to produce flow images \textbf{(d)}.
	    As an option, \textbf{(c)} a more accurate optical flow algorithm 
	    \cite{Brox-2004} can be used % to incrementally process  optical flow images %(for non real-time setting).
	    (although not in real time).
	    %which can be provided directly as input in place of real-time flow images. %in an offline modality.
	    \textbf{(e)}~RGB and OF images are fed to two separate SSD detection~\cite{liu15ssd} networks (\S~\ref{subsec:intg_det_net}).
	    \textbf{(f)}~Each network outputs a set of detection boxes along with their class-specific confidence scores (\S~\ref{subsec:intg_det_net}).
	    \textbf{(g)}~Appearance and flow detections are fused (\S~\ref{subsec:fs_spatil_flow}). 
	    %and linked up to generate class-specific action tubes %spanning the whole video     
	    %for either \textbf{(e)}~online action prediction and localisation     or \textbf{(f)}~offline spatiotemporal action localisation.     
	    %\textbf{(g)}~In the latter case action paths are temporally trimmed to form the final action tubes (\S~\ref{subsec:action_tube}).
	    Finally \textbf{(h)}, multiple action tubes are built up in an online fashion 
	    by associating current detections with partial tubes (\S~\ref{subsec:action_tube}).
        }
    }
 }
  \label{fig:algorithmOverview} \vspace{-3mm}
\end{figure*}

In this work, we propose an online framework, outlined in Figure~\ref{fig:algorithmOverview}, which overcomes all the above limitations.
The pipeline 
takes advantage of the more recent SSD (Single Shot MultiBox Detector) object detector~\cite{liu15ssd} 
to address issues with accuracy and speed at frame level.
This is possible as SSD eliminates the region proposal generation step and is single-stage, end-to-end trainable.
 
To leverage the performance of  SSD, we design a novel single pass online tube building method 
%combined with a novel, completely incremental multi-target tube generation algorithm, 
which leads to both superior accuracy 
(compared to~\cite{Weinzaepfel-2015,Saha2016,peng2016eccv}), especially at realistic detection precision, and real-time detection speed.
%Space-time action tubes \cite{Georgia-2015a,Saha2016} are formed by 
%linking SSD-generated frame-level detection windows over time, namely
%by looking at any time instant for associations between the set of current frame detection boxes 
%and the existing action tubes. % generated up to that time instant.
Unlike previous tube-generation approaches  \cite{Georgia-2015a,Saha2016,peng2016eccv,Weinzaepfel-2015}, our algorithm works in an online fashion as tubes 
are updated frame by frame, together with their overall action-specific scores and labels.
As soon as non-real-time optical flow ~\cite{Brox-2004} is replaced by the
less accurate (but real-time) optical flow algorithm \cite{kroeger2016fast}, the resulting system performs in real time (28fps), with just a little performance degradation, an 
essential feature for real-world applications.

\iffalse
Further, we show that we can replace non-real-time optical flow ~\cite{Brox-2004} with 
less accurate but real-time optical flow proposed by \cite{kroeger2016fast} 
in two stream setting (shown in ~\ref{fig:algorithmOverview}) without sacrificing much performance.
Although, network trained on real-time optical flow~\cite{kroeger2016fast} does not lead comparable performance to 
the network trained more accurate optical flow~\cite{Brox-2004} by itself, but when used in two stream setting 
it can provide similar boost as provided by accurate optical flow~\cite{Brox-2004} network in parallel to appearance stream.
\fi

%%% %%%%%%REMOVED BY GUR , THESE ARE DETAILS
%When optical flow is computed in real time from the input RGB frames~\cite{kroeger2016fast},
%the system achieves a test time detection speed of 30 fps (frames per second).
%If only appearance (RGB inputs) is considered, the framework performs at a remarkable 52 fps on 
%a common machine with Intel Xeon CPU@2.80GHz (8 cores), and one NVIDIA Titan X GPU. 

%\MS{It is not because it is real-time that prediction is possible, this doesn't make sense.}
The incremental nature of our system makes it possible to accurately foresee the class label of an entire test video 
and localise action instances within it by just observing a small fraction of frames (\emph{early action prediction and localisation}).
Such a system has been recently proposed by
Soomro~\etal~\cite{Soomrocvpr2016}, who showed that both action prediction and online localisation performance % is this temporal localisation? Yes both both S/T
improve over time as more and more video frames become available.
Using \cite{Soomrocvpr2016} as a baseline, we report here new state-of-the-art results
%on both early action prediction and online action localisation 
on the temporally trimmed J-HMDB-21 videos.
%\emph{From real time detection to action prediction}.
%In this paper, however, we go even beyond real time recognition to demonstrate the early action label 
%prediction capabilities of our framework, in a setting where predictions are formulated based on 
%the partially observed action labels (i.e. observed from a small subset of the entire video sequence).
%partially observed action instances.
Furthermore, compared to %Soomro~\etal~
\cite{Soomrocvpr2016}, we are able to
demonstrate action prediction and localisation capabilities 
from partially observed \emph{untrimmed} streaming videos on the challenging UCF101-24 dataset,
while retaining real-time detection speeds. 

\textbf{Contributions.} In summary, 
we present a holistic framework for the real-time, online spatial and temporal localisation of multiple action instances in videos which:

\begin{enumerate}[leftmargin=0.5cm,noitemsep,nolistsep]
\item incorporates the newest SSD~\cite{liu15ssd} neural architecture %of a very recent real-time object detector
to predict frame-level detection boxes and the associated action class-specific
confidence scores, in a single-stage regression and classification approach (\S~\ref{subsec:intg_det_net});
\item devises an original, greedy algorithm capable of generating multiple action tubes incrementally (\S~\ref{subsec:action_tube});
%(see Fig.~\ref{fig:algorithmOverview} (g)), 
\item 
%as a result, can 
provides early action class label predictions %(\S~\ref{subsec:action_prediction}) 
and online spatio-temporal localisation results (Fig. 1) %\ref{fig:introductionTeaser}) 
from partially observed action instances in untrimmed videos;%

\item functions in real-time, while outperforming the previous (offline) state of the art 
on the untrimmed videos of UCF101-24 dataset.% and paying a small performance drop on J-HMDB-21.
%by combining the CNN based frame-level detections and an efficient Viterbi forward-backward algorithm to generate incremental action paths~\cite{bmvc2016_paper}.
\end{enumerate}
%% SHORTEN a LITTLE
To the best of our knowledge, our framework is the first with a demonstrated ability to perform online spatial and temporal action localisation.
%While contribution might seem incremental when looked at independently, but 
An extensive empirical evaluation
% our complete framework on benchmarks %major action detection datasets %(UCF101 and J-HMDB-21)
demonstrates that our approach:
\begin{itemize}[leftmargin=0.5cm,noitemsep,nolistsep]
\item significantly outperforms current offline methods, especially on realistic detection thresholds of $0.5$ or greater;% (\S~\ref{sec:online-st-localisation});
\item is capable of superior early action prediction performance compared to the state of the art~\cite{Soomrocvpr2016};
\item achieves a real-time detection speed (upto 40fps), that is 5 to 6 times faster than previous works (\S~\ref{sec:detection-speed}).
\iffalse
of 30 frames per second when computing optical flow in real time \cite{kroeger2016fast}, with only one frame latency,
and is even faster (52fps) when processing mere RGB images
\fi
%DONEi confirm rate one frame latency becuae we use parllel thread to compute flow.
\end{itemize}

\noindent
Our code is available online at \url{https://github.com/gurkirt/realtime-action-detection}.

\section{Related work}

%Recently, inspired by the record-breaking performance of CNNs in image classification \cite{krizhevsky2012}  and object detection from images \cite{girshick-2014},
Deep learning architectures have been increasingly applied of late to action classification \cite{Shuiwang-2013,Karpathy-2014,Simonyan-2014,tran2014learning}, 
spatial \cite{Georgia-2015a}, temporal~\cite{Shou2016} and spatio-temporal~\cite{Weinzaepfel-2015}
action localisation, and event detection \cite{xu2014discriminative}. 
%Nevertheless, other ways of tackling these problems have been recently proposed.

% Tran \etal~\cite{tran2014learning} used three dimensional CNNs to learn spatiotemporal features and applied their model to solve several 
% vision problems such as action, scene and object classification, achieving comparable classification 
% accuracy on very large scale datasets like UCF-101~\cite{?},Sport1M~\cite{?}. 
%%-------------------------------------------
%%--- spatial and temproal sperataly 
%%--------------------------------------------

\emph{Spatial action localisation} 
%problem is catching the interest of researchers and it 
is typically addressed using segmentation~\cite{Lu_2015_CVPR,Soomro2015,jain2014supervoxel}
or region proposal and actionness~\cite{Georgia-2015a,wangcvpr2016} -based approaches.
%Supervoxel approaches~\cite{Soomro2015} appear promising, but
%they require substantial computing time and resources, restricting their applications to offline settings. 
Gkioxari and Malik \cite{Georgia-2015a}, in particular, have built on \cite{girshick-2014} and \cite{Simonyan-2014} 
to tackle spatial action localisation in temporally trimmed videos only,
using Selective-Search region proposals, fine-tuned CNN features and a set of one-vs-rest SVMs. 
These approaches are restricted to trimmed videos. 
%Furthermore, they fail in scenarios involving action instances overlapping in space and time, 
%or action instances with vast temporal-extent variance ~\cite{jain2014supervoxel}.
%spatial action localisation

\emph{Temporal action detection} is mostly tackled using expensive sliding window 
\cite{laptev2007retrieving,gaidon2013temporal,tian2013spatiotemporal,oneata2014efficient,wang2015robust} 
approaches. 
%The latter deliver good results~\cite{oneata2014efficient,gaidon2013temporal,tian2013spatiotemporal}, 
%but they are too inefficient to work in real-time.
%The recent efforts in this area are \cite{Shou2016,yeungcvpr2016,yeung2015every}.
Recently, deep learning-based methods have led to significant advances.
For instance, Shou~\etal~\cite{Shou2016} have employed 3D CNNs~\cite{Shuiwang-2013,tran2014learning} to
address temporal action detection in long videos. 
LSTMs are also increasingly being used~\cite{yeung2015every,de2016online,singhmulti,yeungcvpr2016} to address the problem.
Dynamic programming has been employed to solve the problem efficiently 
\cite{kulkarni2015continuous,Evangel-2014,singh2016untrimmed}.
Some of the above works~\cite{yeung2015every,de2016online,Evangel-2014,yeung2015every} can perform action detection in an online fashion.  
However, unlike our framework, all these methods
%\cite{Shou2016,yeungcvpr2016,wang2014video,wang2015robust,laptev2007retrieving,kulkarni2015continuous,Evangel-2014} 
only address {temporal}, as opposed to spatial and temporal, action detection.

\emph{Spatio-temporal action localisation} can be approached in a supervised~\cite{peng2016eccv,Saha2016},
semi-supervised \cite{vanGemert2015apt,Weinzaepfel-2015}, or weakly supervised~\cite{sapienza2014,Sultani2016} manner.
Inspired by Oneata~\etal~\cite{oneata2014efficient} and Jain~\etal~\cite{jain2014tublet}, %jain2014tublet,vanGemert2015apt,
Gemert~\etal~\cite{vanGemert2015apt} use unsupervised clustering to generate a small set of bounding box-like spatio-temporal action proposals. 
As their method is based on dense-trajectory features~\cite{wang-2011}, it fails to detect actions characterised by small motions~\cite{vanGemert2015apt}.
%can be addressed with 
% frame-level region proposal-based approaches~\cite{Georgia-2015a,Weinzaepfel-2015,vanGemert2015apt,bmvc2016_paper,chen2015action,yucvpr2015}, 
% which has come to fore front of the problem.
Weinzaepfel~\etal's work \cite{Weinzaepfel-2015} performs both temporal and
spatial detections by coupling frame-level EdgeBoxes \cite{zitnick2014edge} region proposals 
with a tracking-by-detection framework. 
%based on a novel track-level descriptor called a Spatio-Temporal Motion Histogram. 
However,  temporal trimming is still achieved via a multi-scale sliding window over each track, making the approach inefficient for longer video sequences.
More recently, Saha \etal \cite{Saha2016} and Peng \etal \cite{peng2016eccv} have made use of supervised region proposal networks (RPNs) \cite{ren2015faster} 
to generate region proposal for actions on frame level, and solved the S/T association problem 
via 2 recursive passes over frame level detections for the entire video by dynamic programming.
Using a non real-time and 2-pass tube generation approach, however, makes their methods offline and inefficient. 
%In addition to that \cite{Saha2016} and \etal \cite{peng2016eccv} make use of non real-time frame level detector faster-rcnn~\cite{ren2015faster}.
%Unlike our approach, however, these methods are not incremental, they rely on two separate stages for region proposal generation and classification, 
%and suffer from slow network speed and heavy optical flow computation. 
In opposition, our framework employs a real-time OF algorithm \cite{kroeger2016fast} and 
a single shot SSD detector \cite{liu15ssd} to build multiple action tubes in a fully incremental way, and in real time.

\emph{Real-time methods}.
Relatively few efforts have been directed at 
simultaneous real time action detection and classification.
Zhang~\etal~\cite{zhangcvpr2016}, for example, accelerate the two-stream CNN architecture of~\cite{Simonyan-2014}, 
performing action classification at 400 frames per second.
Unlike our method, however, theirs cannot perform spatial localisation. 
Yu~\etal~\cite{yu2010bmvc} evaluate their real-time continuous action classification approach 
on the relatively simpler KTH~\cite{schuldt2004recognizing} and 
UT-interaction~\cite{ryoo2010ut} datasets.  % FABIO: DO THEY DO LOCALISATION AS WELL?
To the best of our knowledge, this is the first work to address real-time action localisation.
%They report a real-time evaluation of their method for action classification only with the speed of 17 frames per second,
 %whereas we address in real time the more complex spatio-temporal action localisation (and classification) problem.
  % yes, but what is their approach?
%In this work, instead, we report an extensive evaluation of our proposed real-time  action prediction and localisation 
%method on relatively large scale datasets: UCF-101~\cite{soomro-2012}, J-HMDB-21~\cite{J-HMDB-Jhuang-2013}
%and on a highly complex and fine grained dataset (with multiple and co-occurring actions): LIRIS HARL dataset~\cite{liris-harl-2012}.
%%-------------------------------------------
%%--- action prediciton  ---
%%--------------------------------------------
% ACTION 7
% lacking structure and conclusions, we need contrast say generative approaches (which I think are dominant) with what we do
%Researchers are starting to show interest in early event detection \cite{Earlyryoo2011human} in temporal trimmed videos 
%and future action prediction \cite{yeung2015every,yeungcvpr2016} in untrimmed videos as the natural logical step after real time detection. 
%along with spatio-temporal action detection.

\emph{Online action prediction}. 
Early, online action prediction has been studied using dynamic bag of words \cite{Earlyryoo2011human}, structured SVMs \cite{hoai2014max}, 
hierarchical representations \cite{lan2014hierarchical}, LSTMs and Fisher vectors \cite{de2016online}.
Once again, 
unlike our framework, these approaches \cite{Earlyryoo2011human,hoai2014max,lan2014hierarchical} do not perform online action localisation. 
%FABIO: WE COULD STILL USE THEM AS BASELINE FOR THE PREDICTION BIT, RIGHT? 
Soomro~\etal~\cite{Soomrocvpr2016} recently proposed an online method which can predict an action's label 
and location by observing a relatively smaller portion of the entire video sequence.
%Their method improves detection accuracy over time. 
%And, they use segmentation to perform online prediction with the help of SVM model trained on fixed length segments of training videos.
However, \cite{Soomrocvpr2016} only works on temporally trimmed videos and not in real-time, due to expensive segmentation.

\section{Methodology} \label{sec:methodology}
%\MS{Can be shortened or removed in an emergency.}
%\MS{Remove extra braces around sub-figure letters, or make consistent throught.}
As outlined in Fig.~\ref{fig:algorithmOverview}, our approach exploits an integrated detection
network~\cite{liu15ssd} (\S~\ref{subsec:intg_det_net}-Fig.~\ref{fig:algorithmOverview}e)
to predict detection boxes and class-specific confidence scores for appearance and flow (\S~\ref{subsec:optical-flow}) video frames independently.
One of two alternative fusion strategies (\S~\ref{subsec:fs_spatil_flow}-Fig.~\ref{fig:algorithmOverview}g)
is then applied.
%to fuse the cues extracted from appearance and flow-based detection networks.
%These detection boxes are fused (\S~\ref{subsec:fs_spatil_flow}-Fig.~\ref{fig:algorithmOverview}(d)) to generate a new set of boxes with associated scores for each frame of the video. 
Finally, action tubes are built incrementally in an online fashion and in real time, using a new efficient action tube generation algorithm
 (\S~\ref{subsec:action_tube}-Fig.~\ref{fig:algorithmOverview}h),
 which can be applied to early action prediction (\S~\ref{subsec:action_prediction}).
%generated in an online fashion using a single pass greedy strategy (\S~\ref{subsec:act_pred}, Fig.~\ref{fig:algorithmOverview}(e)),
% either in offline mode by solving a cascade of two optimisation problems via dynamic programming as 
%in \cite{Saha2016} (\S~\ref{subsec:action_tube}, Fig.~\ref{fig:algorithmOverview}(f) and (g)), or 
%All elements are described below.

\subsection{Optical flow computation} \label{subsec:optical-flow}

The input to our framework is a sequence of RGB images. 
As in prior work in action localisation~\cite{Saha2016,Georgia-2015a,Weinzaepfel-2015},
we use a two-stream CNN approach \cite{Simonyan-2014} in which optical flow and appearance are processed in two parallel, distinct streams.
As our aim is to perform action localisation in real-time, 
we employ real-time optical flow (Fig.~\ref{fig:algorithmOverview}b)
\cite{kroeger2016fast} to generate the flow images (Fig.~\ref{fig:algorithmOverview}d).
As an option, one can compute optical flow more accurately (Fig.~\ref{fig:algorithmOverview}c),
using Brox \etal's \cite{Brox-2004} method.
We thus train two different networks for the two OF algorithms, 
while at test time only one network is used 
%depending on the choice of either real-time or non real-time detection.
depending on whether the focus is on speed rather than accuracy.
Following the transfer learning approach on motion vectors of \cite{zhangcvpr2016}, 
we first train the SSD network on accurate flow results, 
to later transfer the learned weights to initialise those of the real time OF network. 
Performance would degrade whenever transfer learning was not used.

\subsection{Integrated detection network} \label{subsec:intg_det_net}
We use a single-stage convolutional neural network (Fig.~\ref{fig:algorithmOverview}e) for bounding box prediction and classification,
which follows an end-to-end trainable architecture proposed in~\cite{liu15ssd}.
The architecture unifies a number of functionalities in single CNN 
which are, in other action and object detectors, performed
by separate components~\cite{Georgia-2015a,Weinzaepfel-2015,ren2015faster,Saha2016}, namely: 
{(i)} region proposal generation, 
{(ii)} bounding box prediction and 
{(iii)} estimation of class-specific confidence scores for the predicted boxes.
This allows for relatively faster training and higher test time detection speeds.

\textbf{Detection network design and training.}
{For our integrated detection network we adopt the network design and architecture of the SSD~\cite{liu15ssd} object detector, 
with an input image size of $300\times300$.
We do not use the $512\times512$ SSD architecture~\cite{liu15ssd}, as detection speed is much slower~\cite{liu15ssd}.
As in~\cite{liu15ssd}, we also use an ImageNet pretrained VGG16 network 
provided by~\cite{liu15ssd} (\url{https://gist.github.com/weiliu89/2ed6e13bfd5b57cf81d6}).
We adopt the training procedure described by ~\cite{liu15ssd} along with their
publicly available code for network training (\url{https://github.com/weiliu89/caffe/tree/ssd}).
We use a learning rate of $0.0005$ for the appearance stream and of $0.0001$ for the flow stream on UCF101-24, 
whereas that for JHMDB is set to $0.0001$ for both appearance and flow. 
All implementation details are in the supplementary material.}

\subsection{Fusion of appearance and flow cues}\label{subsec:fs_spatil_flow}
%%%%%%% NO CHNAGE
The detection boxes generated by the appearance and flow detection networks (Fig.~\ref{fig:algorithmOverview}f) 
need to be merged to improve robustness and accuracy (Fig.~\ref{fig:algorithmOverview}g).
We conducted experiments using two distinct fusion strategies.\\
\textbf{Boost-fusion}. Here we follow the approach in~\cite{Saha2016}, with a minor modification.
Firstly, we perform L-1 normalisation on  the detection boxes' scores after fusion.
Secondly, we retain any flow detection boxes for which an associated appearance based box was not found,
as we found that discarding the boxes lowers the overall recall.\\
\textbf{Fusion by taking the union-set.}\label{subsec:fusion_by_union}
 A different, effective fusion strategy consists in retaining the union  
$\{b^{a}_i\}\cup\{b^{f}_j\}$ of the two sets of appearance $\{b^{a}_i\}$ and flow $\{b^{f}_j\}$ detection boxes, respectively.
The rationale is that in UCF-101, for instance, several action classes
(such as `Biking', `IceDancing', or `SalsaSpin') have concurrent action
instances in the majority of %train and test 
video clips: an increased number of detection boxes may so
help to localise concurrent action instances.

\subsection{Online action tube generation} \label{subsec:action_tube}

%\subsubsection{Problem formulation}

Given a set of detections at time $t=1..T$, for each given action class $c$, we seek the sets of consecutive detections (or \emph{action tubes}) $\mathcal{T}_c = \{ b_{t_s},…,b_{t_e} \}$ which, among all possible such collections, are more likely to constitute an action instance. This is done separately for each class, so that results for class $c$ do not influence those for other classes.
We allow the number of tubes $n_c(t)$ to vary in time, within the constraint given by the number of available input detections. We allow action tubes to start or end at any given time.
Finally, we require: (i) consecutive detections part of an action tube to have spatial overlap above a threshold $\lambda$; (ii) each class-specific detection to belong to a single action tube; (iii) the online update of the tubes' temporal labels.
\\
Previous approaches to the problem \cite{Georgia-2015a,Saha2016} constrain tubes to span the entire video duration. In both \cite{Saha2016} and \cite{peng2016eccv}, in addition, action paths are temporally trimmed to proper action tubes using a second pass of dynamic programming.

\iffalse
{We define an \emph{action tube} as a connected sequence of detection boxes in time, without interruptions,
associated with a same action class $c$, 
starting and ending at arbitrary time points $t_s$, $t_e$ 
in the video: $\mathcal{T}_c = \{ {b}_{t_{s}}, ... , {b}_{t_{e}}\}$.
Tubes should not be constrained to span the entire video duration~\cite{Saha2016}, unlike in ~\cite{Georgia-2015a}.
Saha \etal ~\cite{Saha2016} solves this problem using two passes.
First they follow ~\cite{Georgia-2015a} to recursively construct multiple action paths for a video, spanning the entire video duration~\cite{Saha2016}. Then,
action paths are temporally trimmed to proper action tubes using a second pass of dynamic programming.
Peng \etal~\cite{peng2016eccv} also use a similar two pass approach to tube generation.}
%\MS{This sounds like it should be in the related work section.}
\fi

In opposition, we propose a simple but efficient online action tube generation 
algorithm which incrementally (frame by frame) builds multiple action tubes for each action class in parallel.  
Action tubes are treated as `tracklets', as in multi-target tracking approaches~\cite{Milan:2016:PAMI}. 
We propose a greedy algorithm (\ref{subsub:algo}) similar to \cite{majecka2009statistical,singh2010categorising} 
for associating detection boxes in the upcoming frame with the current set of (partial) action tubes. 
Concurrently, each tube is temporally trimmed in an online temporal labelling (\ref{subsub:temp_label}) setting.

%While cannot claim as yet that our algorithm is unique in meeting them, the above constraints rule out a number of options.

\subsubsection{A novel greedy algorithm}\label{subsub:algo}
The input to the algorithm is the fused frame-level detection boxes
with their class specific scores (Sec. \ref{subsec:fs_spatil_flow}). % at each time step $t$.
{At each time step $t$, the top $n$ class-specific detection boxes $\{b_{c}\}$ are selected by
applying non-maximum suppression on a per-class basis.
%At each time point non-maximal suppression is applied per class $c$ and top $n$ boxes are selected per class.
%We start building class-specific action tubes by selecting top $n$ detection boxes from the very first incoming video frame.
%The who process is described in steps below.
At the first frame of the video, $n_c(1) = n$ action tubes per class $c$ are initialised using the $n$ detection boxes at $t=1$.
The algorithm incrementally grows the tubes over time by adding one box at a time.
The number of tubes $n_c(t)$ varies with time, as new tubes are added and/or old tubes are terminated.

At each time step, we sort the existing partial tubes so that the best tube can potentially match the best box from the set of detection boxes in the next frame $t$.
Also, for each partial tube $\mathcal{T}_{c}^i$ at time $t-1$, we restrict the potential matches to detection boxes at time $t$ whose IoU (Intersection over Union) 
with the last box of $\mathcal{T}_{c}^i$ is above a threshold $\lambda$.
In this way tubes cannot simply drift off, and they can be terminated 
%This is done so tube doesn't drift off and could also terminate 
if no matches are found for $k$ consecutive frames. 
Finally, each newly updated tube is temporally trimmed by
performing a binary labelling %\mbox{${l} \in \{c,0\}$} for each tube at each time step after associating a box with a tube by 
using an online Viterbi algorithm. This is described in detail in Sec. \ref{subsub:temp_label}.

Summarising, 
action tubes are constructed by applying the following 7 steps to every new frame at time~$t$:}
% [SUMAN] I am commenting out the following for now
%--------------------------------------------------------------------------------------------------------------------------------------------------------
% Action tubes are generated for each class individually after the applying non-maximal suppression per class and maximum of top $n$ boxes are kept for each class.
% At the start of video, we initialise $n$ action tubes from top $n$ detection boxes form first frame. 
%For every next frame, take these 6 steps, 
%\MS{Put as an Algorithm and refere to steps from the text, for instance temporal binary labelling is not clear when reading it in this order.}
\begin{enumerate} [topsep=0pt,itemsep=-1ex,partopsep=1ex,parsep=1ex]
 \item Execute steps 2 to 7 for each class $c$.
 \item Sort the action tubes generated up to time $t-1$ in decreasing order, based on 
 the mean of the class scores of the tube's member detection boxes.
 \item {\footnotesize \textbf{LOOP START: }} $i = 1 $ to $n_c(t-1)$ - traverse the sorted tube list.
 \item Pick tube $\mathcal{T}_{c}^i$ from the list and find a matching box for it
 among the $n$ class-specific detection boxes $\{b_{c}^j, j=1,...,n\}$ at frame $t$ based on the following conditions:
 \begin{enumerate} [topsep=0pt,itemsep=-1ex,partopsep=1ex,parsep=1ex]
  \item {for all $j=1,...,n$, if the IoU between the last box of tube $\mathcal{T}_c^{i}$ and the detection box $b_{c}^{j}$
   is greater than $\lambda$, then add it to a potential match list $\mathcal{B}^i$;}
 \item if the list of potential matches is not empty, $\mathcal{B}^i \neq \emptyset$, select the box $b_{c}^{max}$ from $\mathcal{B}^i$ 
  with the highest score for class $c$ as the match, and remove it from the set of available detection boxes at time $t$;
 \item if $\mathcal{B}^i = \emptyset$, retain the tube anyway, without adding any new detection box, unless more than $k$ frames have passed with no match found for it.
  %\item select the box out of the potential matches which has highest score for class $c$ and remove that box from detection boxes list.
 \end{enumerate}
 \item Update the temporal labelling %\mbox{${l}_{t} \in \{c,0\}$} 
 for tube $\mathcal{T}_c^{i}$ using the score $s(b_{c}^{max})$ of the selected box $b_{c}^{max}$ (see \S~\ref{subsub:temp_label}).
% \MS{Explain that you keep track of the edges and you don't need to do full viterbi at each iteration.}
 \item {\footnotesize\textbf{LOOP END }}
 %\item Pick the next tube from the sorted list and repeat steps 2) and 4).
 \item If any detection box is left unassigned, start a new tube at time $t$ using this box.
\end{enumerate}
In all our experiments, we set $\lambda = 0.1$, $n = 10$, and $k=5$.

\subsubsection{Temporal labelling}\label{subsub:temp_label}

Although $n$ action tubes per class are initialised at frame $t=1$, we want
all action specific tubes to be allowed to start and end at any arbitrary time points $t_s$ and $t_e$.
The online temporal relabelling step 5. in the above algorithm is designed to take care of this.

Similar to \cite{Saha2016,Evangel-2014}, %each tube is temporally trimmed by assigning 
each detection box $b_r$, $r=1,...,T$ in a tube $\mathcal{T}_c$, 
where $T$ is the current duration of the tube and $r$ is its temporal position within it, is assigned
a binary label \mbox{${l}_{r} \in \{c,0\}$}, 
where $c$ is the tube's class label and $0$ denotes the background class.
The temporal trimming of an action tube thus reduces to finding an optimal binary labelling $\mathbf{l} = \{ l_1,...,l_T \}$ for all the constituting bounding boxes.
This can be achieved by maximising for each tube $\mathcal{T}_c$ the energy:
%\MS{This gives the impression that you solve the entire problem as each frame is added. Consider writing the equation that optimises as each new frame is added.}
%\GUR{Keeping this equation to show exactly what we solving otherwise it will look very adhoc. Tracking of two edge is explained in next Paragraph.}
\begin{equation} \label{eqn:secondpassenergy}
%E(\mathcal{T}_c) =  \sum_{t=1}^T s_{l_t}({b}_t) - \alpha_{l} \sum_{t=2}^T \psi_l \left(\Scalar{l}_{t}, \Scalar{l}_{t-1} \right) ,
E(\mathbf{l}) =  \sum_{r=1}^T s_{l_r}({b}_r) - \alpha_{l} \sum_{r=2}^T \psi_l \left(\Scalar{l}_{r}, \Scalar{l}_{r-1} \right) ,
\end{equation}
where $s_{l_r}(b_r) = s_{c}(b_r)$ if $l_r=c$, $1-s_{c}(b_r)$ if $l_r=0$,
$\alpha_{l}$ is a scalar parameter, and
the pairwise potential $\psi_l$ is defined as: $\psi_l(l_r,l_{r-1}) = 0$ if $l_r = l_{r-1}$, $\psi_l(l_r, l_{r-1}) =\alpha_c$ otherwise.
%We update the labelling costs of each tube at each time step. %(\S~Algorithm~\ref{algo:tube_generation}).
\vskip 0.2cm
\noindent
\textbf{Online Viterbi.}
The maximisation problem~(\ref{eqn:secondpassenergy}) can be solved by Viterbi dynamic programming~\cite{Saha2016}.
{An optimal labelling $\hat{\mathbf{l}}$ for a tube $\mathcal{T}_c$ can be generated 
by a Viterbi backward pass at any arbitrary time instant $t$ in linear time.
We keep track of past box-to-tube associations %\MS{why two? something like: the min cost incoming edges to each node} 
from the start of the tube up to $t-1$,
which eliminates the requirement of an entire backward pass at each time step.
This makes temporal labelling very efficient, and suitable to be used in an online fashion. 
This can be further optimised for much longer videos by finding the coalescence point \cite{vsramek2007line}.
As stated in step 5. above, the {temporal labelling} of each tube is updated at each time step whenever a new box is added.
In the supplementary material, we present a pseudocode of our online action tube generation algorithm.}
%We keep track of the dynamic programming matrix for each tube-specific maximisation problem (\ref{eqn:secondpassenergy}) at each time step $t$, by paying a little memory cost which then facilitates backtracing in constant time. 

\subsection{Early action prediction} \label{subsec:action_prediction}
%Our online tube generation algorithm (\S \ref{subsec:action_tube}) can be used to predict the label of an incoming video at any time instant $t$.
As for each test video multiple tubes are built incrementally at each time step $t$ (\S \ref{subsec:action_tube}),
we can predict at any time instant the label of the whole video as the label of the current highest-scoring tube, where
the score of a tube is defined as the mean of the tube boxes' individual detection scores:
%\begin{equation} \label{eq:label-prediction}
$
\hat{c}(t) = \arg \max_{c} \left ( \max_{\mathcal{T}_c} \frac{1}{T}\sum_{r=1}^T s(b_r) \right ).
$
%\end{equation}
%The highest-scoring tube $\hat{\mathcal{T}}_{\hat{c}}$ can also be used to evaluate the online spatio-temporal detection accuracy of the system (\S \ref{sec:online-st-localisation}).
%Label of the best tube make it possible to can used to compute accuracy of our 
%Further, we can evaluate sptio-temporal localisation at any time point. 
%Each video has multiple tubes being constructed for each class in parallel,
%we select the label tube with highest score to label the video at any time. 
%At any given time point, we can label an input video with the label of the highest scored tube
%Tubes score is defined by mean of the scores so far.
%We show evaluation of our method in \ref{sec:prediction} 
%We incrementally extract the action tubes at 10 different time points, starting at $10\%$ video frames and with step size of $10\%$ of video frames.
%We take the label of the video from the tube which has best classification score.

\section{Experiments} \label{subsec:discussion}
We test our online framework (\S~\ref{sec:methodology}) on two separate challenging  
problems: 
i) early action prediction (\S~\ref{sec:prediction}),
ii) online spatio-temporal action localisation (\S~\ref{sec:online-st-localisation}),
including a comparison to offline action detection methods.
\iffalse
iii) a comparison with offline action detection methods (\S~\ref{sec:Spatio-temporal}).
\fi
Evidence of real time capability is provided in (\S~\ref{sec:detection-speed}).

In all settings we generate results by running our framework in five different `modes':
1) \emph{Appearance (A)} -- only RGB video frames are processed by a single SSD;
2) \emph{Real-time flow (RTF)} -- optical flow images are computed in real-time~\cite{kroeger2016fast} and fed to a single SSD;
3) \emph{A$+$RTF}: both RGB and real-time optical flow images are processed by a SSD in two separate streams;
4) \emph{Accurate flow (AF)} optical flow images are computed as in~\cite{Brox-2004}, and
5) \emph{A$+$AF}: both RGB and non real-time optical flow frames \cite{Brox-2004} are used.
\\
Modes 1), 2) and 3) run in real-time whereas modes 4) and 5)'s performances are non real-time (while still working incrementally), 
due to the relatively higher computation time needed to generate accurate optical flow.
\iffalse
Note that all experiments are conducted in a fully online fashion, using the proposed tube generation algorithm (\S~\ref{subsec:action_tube}).
\fi

\textbf{Datasets.}
{We evaluate our model on the UCF-101-24~\cite{soomro-2012} and 
J-HMDB-21~\cite{J-HMDB-Jhuang-2013} benchmarks.
\textbf{UCF101-24} is a subset of UCF101~\cite{soomro-2012}, 
one of the largest and most diversified and challenging action datasets. 
Although each video only contains a single action category, 
it may contain multiple action instances (upto 12 in a video) of the same action class, 
with different spatial and temporal boundaries.
A subset of 24 classes out of 101 comes with spatio-temporal localisation annotation, 
released as bounding box annotations of humans with THUMOS-2013 challenge\footnote{http://crcv.ucf.edu/ICCV13-Action-Workshop/download.html}.
On average there are 1.5 action instances per video, each action instance covering 70\% of the duration of the video. 
For some classes, instances avergae duration can be as low as 30\%.
%\footnote{More details about the untrimmed nature of UCF101-24 videos are given in the supplementary material.}
%\footnote{http://crcv.ucf.edu/ICCV13-Action-Workshop/index.files/UCF101_24Action_Detection_Annotations.zip}.
As in previous spatio-temporal action detection works \cite{Saha2016,yu2015fast,peng2016eccv,Weinzaepfel-2015}, 
we test our method on split 1. 
\textbf{J-HMDB-21}~\cite{J-HMDB-Jhuang-2013}  
is a subset of the HMDB-51 dataset~\cite{HMDBkuehne2011hmdb} with 21 action categories and 928 videos, 
each containing a single action instance and trimmed to the action's duration.}

Note that the THUMOS~\cite{gorban2015thumos} and ActivityNet~\cite{caba2015activitynet} datasets are not suitable for spatiotemporal localisation, as they lack bounding box annotation.

\textbf{Evaluation metrics.}
For the early action label prediction (\S~\ref{sec:prediction}) and 
the online action localisation (\S~\ref{sec:online-st-localisation}) tasks we follow
the experimental setup of~\cite{Soomrocvpr2016}, and use  the
traditional localisation metrics AUC (area under the curve) and mAP (mean average precision).
We report performance as a function of \emph{Video Observation Percentage}, i.e.,
with respect to the portion (\%) of the entire video observed before predicting action label and
location.
We also report a performance comparison to offline methods~\cite{Saha2016,yu2015fast,peng2016eccv,Weinzaepfel-2015} 
using the protocol by Weinzaepfel~\etal~\cite{Weinzaepfel-2015}.

\begin{figure}[t]
  \centering
  \vspace{-2mm}
  \includegraphics[width=0.48\textwidth]{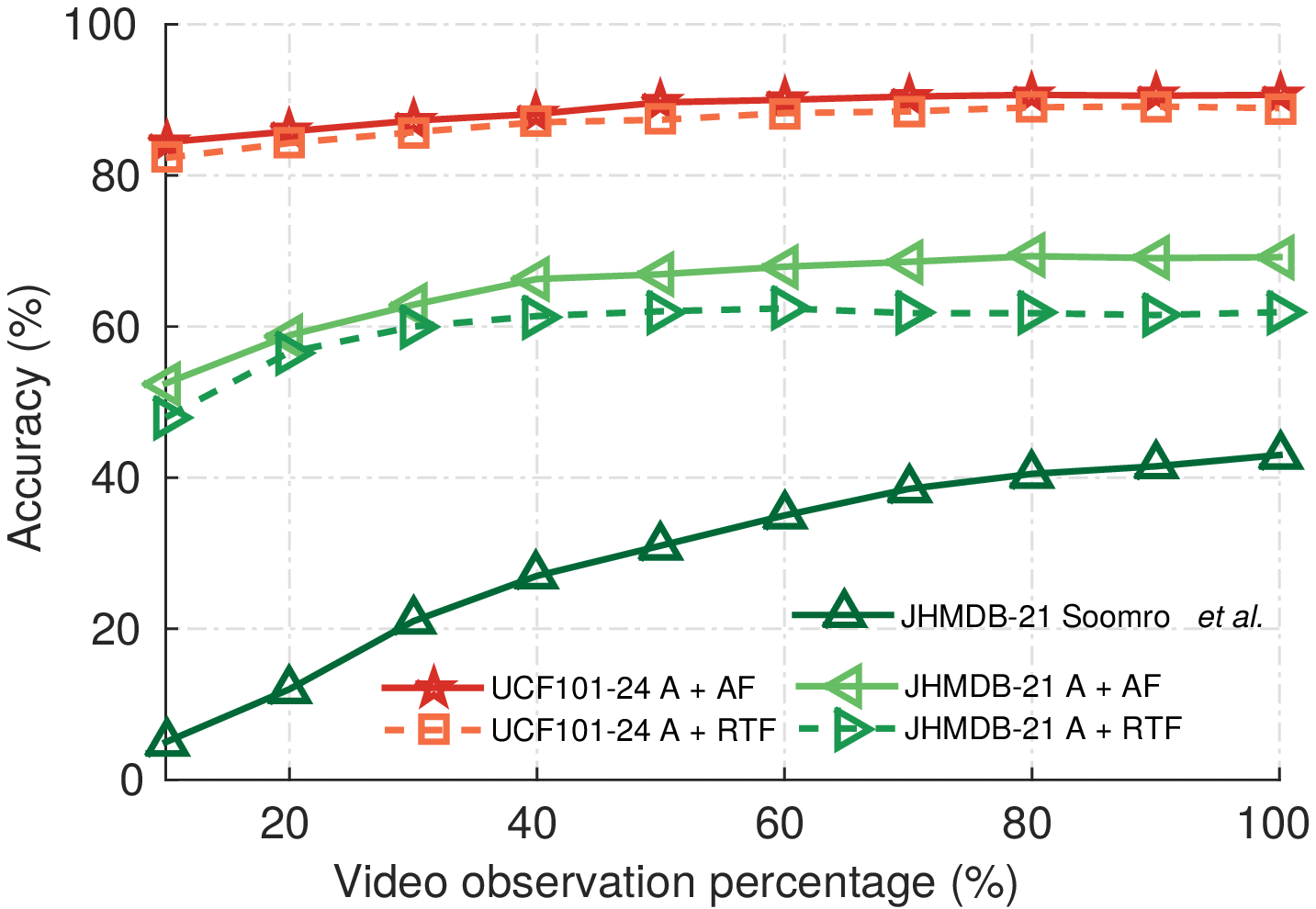}  
  \caption{
    {\small
      \textit{
        Early action label prediction results (accuracy \%) on the UCF101-24 and J-HMDB-21 datasets.
        %; A $+$ AF - Appearance and Accurate Flow; A $+$ RTF - Appearance and
        %real-time optical flow. % on the UCF101 and JHMDB datasets.        
      }
    }
  }
  \label{fig:prediction} \vspace{-3mm}
\end{figure}

\subsection{Early action label prediction} \label{sec:prediction}
Although action tubes are computed by our framework frame by frame, we sample them
at 10 different time `check-points' along each video,
starting at $10\%$ of the total number of video frames and with a step size of $10\%$.
%At each check-point we label a video as per the label of the current tube with 
%the best classification score as explained earlier~(\S~\ref{subsec:action_prediction}).
We use the union-set and boost fusion strategies~(\S~\ref{subsec:fs_spatil_flow}) for UCF101-24 and J-HMDB-21, respectively.
%We also evaluate our pipeline for online action localisation and early action label prediction on J-HMDB-21 and UCF101-24
%and compare the performance with the most recent online action prediction approach~\cite{Soomrocvpr2016} on J-HMDB-21.
%the online action localisation and prediction capabilities of our approach on J-HMDB-21 and UCF101-24, 
%and compared them with that of the  most recent
%online action prediction approach~\cite{Soomrocvpr2016} on J-HMDB-21 dataset.
%\begin{flushleft}
%\textbf{Online early action prediction experiments}
%\end{flushleft}
%We repeat this for each video at each video observation percentage (VOP), 
%that give us the accuracy on each VOP, it is shown in figure \ref{fig:prediction}.
%\textbf{Action prediction on J-HMDB-21.}
Fig.~\ref{fig:prediction} %\textbf{(a)} 
compares the early action prediction accuracy of our approach with that of~\cite{Soomrocvpr2016}, 
as a function of the portion (\%) of video observed. 
Our method clearly demonstrates superior performance, 
as it is able to predict the actual video label by observing a very small portion of the entire video at a very initial stage. 
For instance,  by observing only the initial $10\%$ of the videos in J-HMDB-21, 
we are able to achieve a prediction accuracy of $48\%$ as compared to $5\%$ by Soomro~\etal~\cite{Soomrocvpr2016}, 
which is in fact higher than the $43\%$ accuracy achieved by \cite{Soomrocvpr2016} after observing the \emph{entire} video.
{We do not run comparisons with the early action prediction work by Ma \etal \cite{ma2016learning} for they only show results on ActivityNet~\cite{caba2015activitynet}, as dataset which has only temporal annotations. The early prediction capability of our approach is a subproduct of its being online, as in~\cite{Soomrocvpr2016}: thus, we only compare ourselves with Soomro~\etal~\cite{Soomrocvpr2016} re early action prediction results.}

Compared to \cite{Soomrocvpr2016} we take one step further, 
and perform early label prediction on the untrimmed videos of UCF101-24 as well (see Fig.~\ref{fig:prediction}).
It can be noted that our method performs much better on UCF101-24 than on J-HMBD-21 at the prediction task.
This relatively higher performance may be attributed to the larger number of training examples, subject to more modes of variations, 
present in UCF101-24, which improves the generalisation ability of the model and prevents it from overfitting.
Interestingly, we can observe that the performances of the real-time (A $+$ RTF) 
and non real-time (A $+$ AF) modalities 
%(almost identical on UCF101 and comparable on J-HMDB-21)
are quite similar, which suggests that accurate optical flow might be not so crucial for action classification on UCF101-24 dataset. 
%It will be interesting to evaluate this using more recent and accurate offline optical flow methods.
\iffalse
which strongly supports our framework's ambition to accurately solve the problem in 
real world situations in which real time speed is required. 
\fi
%To the best of our knowledge, we are the first to report early action label prediction (and online action localisation, 
%see \S~\ref{sec:online-st-localisation}) results on the UCF101-24 dataset. 
%the when more than 50\% of the video is observed.
%Further, we can infer that, our method can be used for online real-time action prediction
%where we need to drive interference before the action finishes,
%Which 
%When only half the video is observed we are able to predict the correct label 58.45\% of the time, as compared to the 30\% achieved by Soomro \etal \cite{Soomrocvpr2016}.\\
% \begin{flushleft}
% \textbf{Online action localisation experiments}\\
% \end{flushleft}
\begin{figure}[t]
  \centering
  \vspace{-2mm}
  \includegraphics[width=0.45\textwidth]{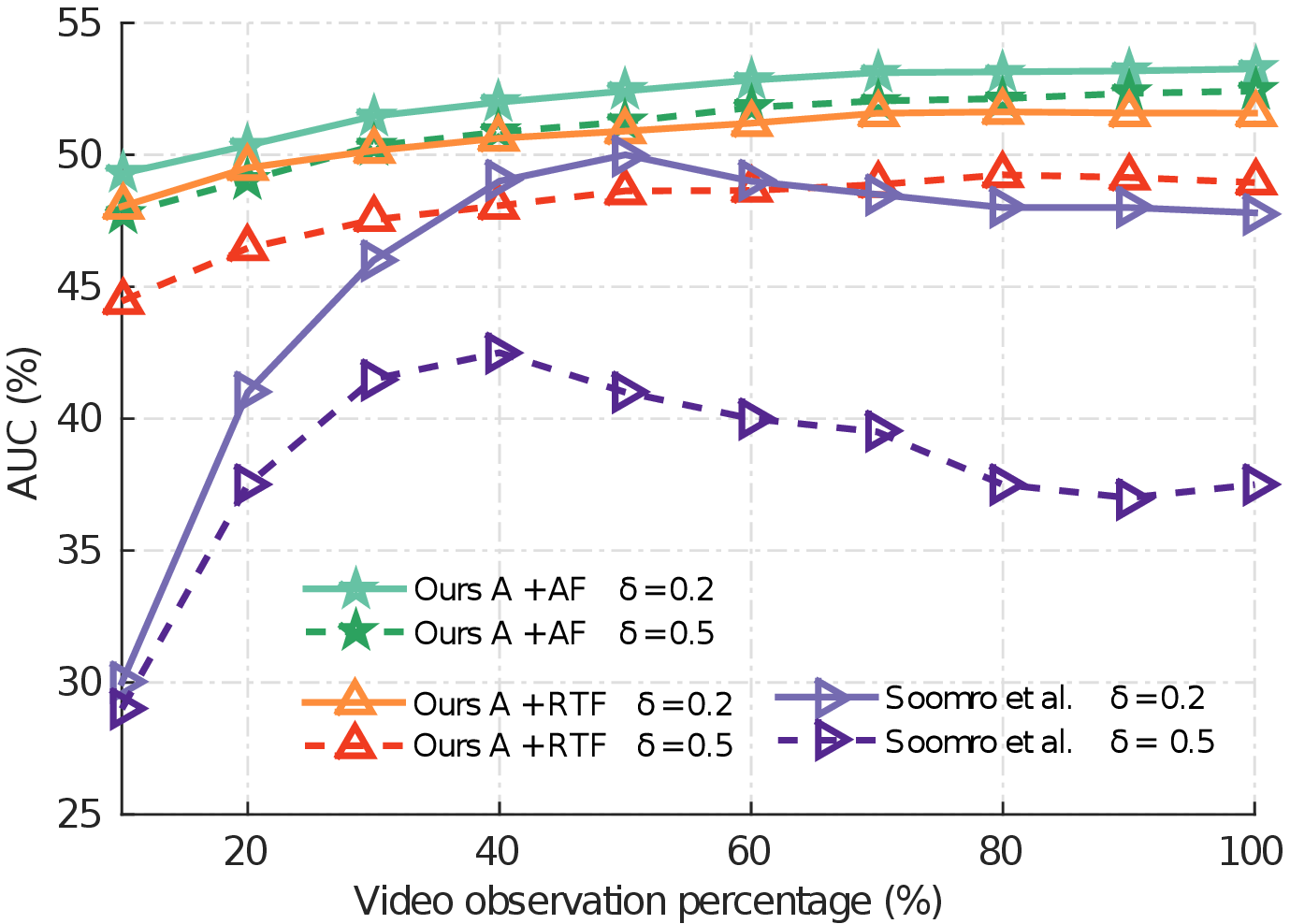}  
  \caption{
    {\small
      \textit{
        Online action localisation results using the AUC (\%) metric on J-HMDB-21, at IoU thresholds of $\delta=0.2$, $0.5$.
        % A $+$ AF - Appearance and Accurate Flow; A $+$ RTF - Appearance and real-time optical flow.
        %We use boost-fusion on J-HMDB-21 dataset. - we have said this at the begining
      }
    }
  }
  \label{fig:localisation_plot_jhmdb} \vspace{-3mm}
\end{figure}
\subsection{Online spatio-temporal action localisation}
\label{sec:online-st-localisation}

\subsubsection{Performance over time}
Our action tubes are built incrementally and carry associated labels and scores at each time step.
At any arbitrary time $t$, we can thus compute the spatio-temporal IoU between the tubes generated by our online algorithm
and the ground truth tubes, up to time $t$.

%We use both the AUC and mean average precision (mAP) metrics to directly 
%compare the online action localisation performance of our method to~\cite{Soomrocvpr2016}.

%\textit{AUC metric}.
Fig.~\ref{fig:localisation_plot_jhmdb} plots the AUC curves 
against the observed portion of the video at different IoU thresholds ($\delta = 0.2$  and $0.5$)
for the proposed approach versus our competitor~\cite{Soomrocvpr2016}.
Our method outperforms~\cite{Soomrocvpr2016} on online action localisation
by a large margin at all the IoU thresholds and video observation percentage.
Notice that our online localisation performance (Fig.~\ref{fig:localisation_plot_jhmdb}) is a stable 
function of the video observation percentage, 
whereas, Soomro~\etal~\cite{Soomrocvpr2016}'s method needs some `warm-up' time to 
reach stability, and its accuracy slightly decreases at the end.
{In addition, \cite{Soomrocvpr2016} only reports online spatial localisation results
on the temporally trimmed J-HMDB-21 test videos, and their approach lacks  
temporal detection capabilities.

Our framework, instead, can perform online spatio-temporal localisation:
to demonstrate this, we present results on the temporally untrimmed UCF101-24 test videos as well. In Fig.~\ref{fig:localisation_plot_mAP} we report
online spatial-temporal localisation results
on UCF101-24 and JHMBD-21 using the standard mAP metric (not reported in \cite{Soomrocvpr2016}).
{Interestingly, for UCF101-24, at a relatively smaller IoU threshold ($\delta = 0.2$) the performance gradually increases over time as more video frames are observed,
whereas at a higher IoU threshold ($\delta = 0.5$) it slightly degrades over time.
A reason for this could be that
UCF101-24 videos are temporally untrimmed and contain multiple action instances, 
so that accurate detection may be challenging at higher detection thresholds (e.g. $\delta = 0.5$).
%most of action instances begin at the start of the video and finish much earlier than end of video.
If temporal labelling is not very accurate, 
as required at high thresholds ($\delta = 0.5$), this might result in more false positives as the video progress, 
hence the observed drop in performance over time. % at $\delta = 0.5$.}

\begin{figure}[t]
  \centering
  \vspace{-2mm}
  \includegraphics[width=0.48\textwidth]{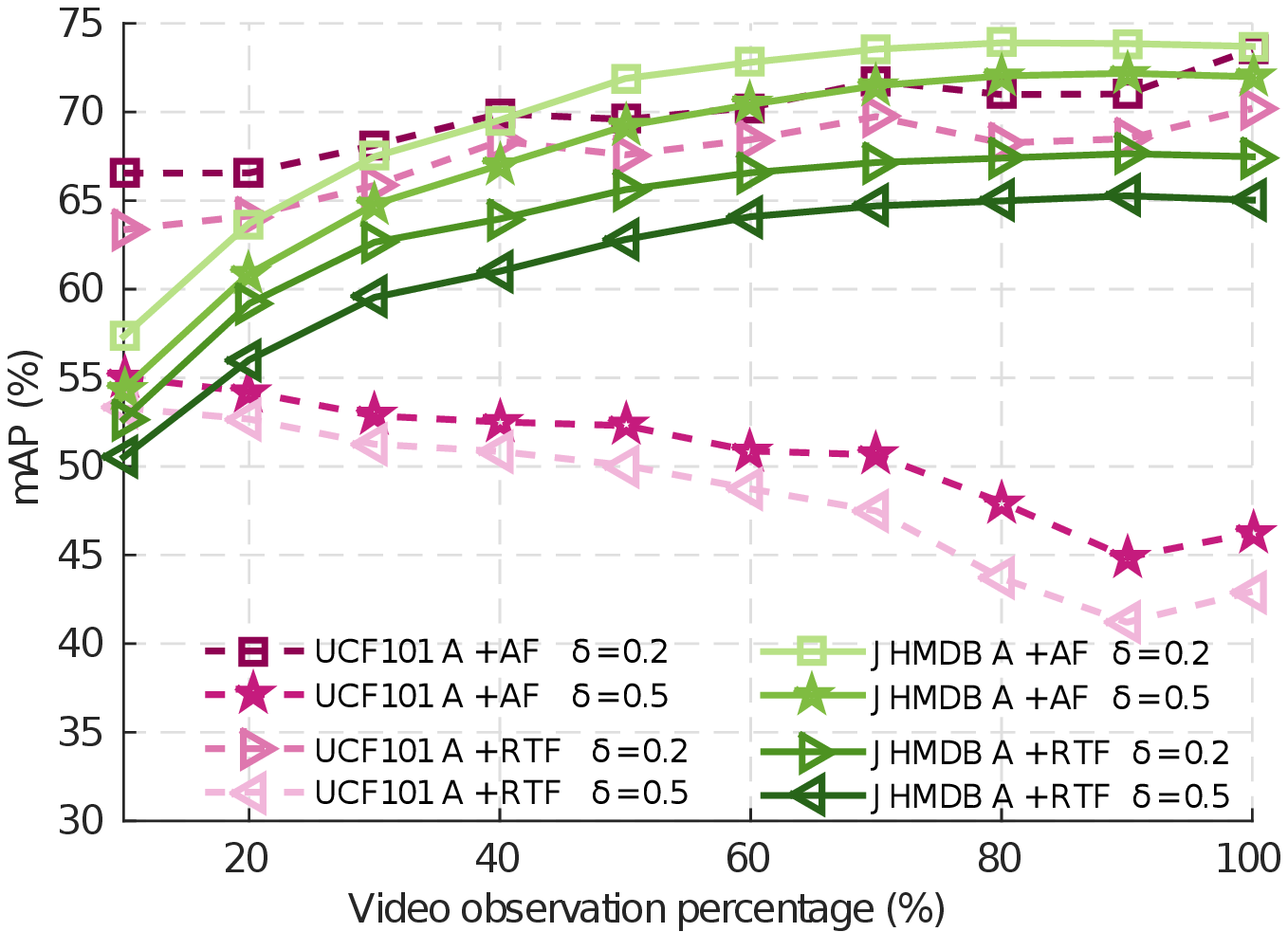}  
  \caption{
    {\small
      \textit{
        Action localisation results using the mAP (\%) metric on UCF101-24 and JHMDB-21, at IoU thresholds of $\delta=0.2,0.5$.
        % A $+$ AF - Appearance and Accurate Flow; A $+$ RTF - Appearance and real-time optical flow.
        %We use union-set fusion strategy for UCF101 and boost-fusion on JHMDB dataset.
      }
    }
  }
  \label{fig:localisation_plot_mAP} \vspace{-5mm}
\end{figure}

\iffalse
% Fabio: add description of internal comparison (b)
\textbf{Comparison of action localisation performance in offline setting.}
% how the local perform varies across time 
Finally, following a similar analysis in Soomro~\etal~\cite{Soomrocvpr2016}'s work, we compare the action localisation
performance of our approach with that of Gkioxari~\etal~\cite{Georgia-2015a} 
and Soomro~\etal~\cite{Soomrocvpr2016} in an offline setting
(i.e., assuming that the system has observed the entire video).
Fig.~\ref{fig:J-HMDB_results_prediction} \textbf{(d)} plots the AUC against different IoU threshold values, showing that
%It is worth noticing that 
our online method outperforms the previously dominating offline approach by 
Gkioxari and Malik~\cite{Georgia-2015a}.
\fi

%\subsection{Spatio-temporal action localisation} \label{sec:Spatio-temporal}
\subsubsection{Global performance}

To demonstrate the strength of our online framework, we compare as well its absolute detection performances to 
those of the top offline competitors~\cite{Saha2016,Weinzaepfel-2015,peng2016eccv,yu2015fast}. 
To ensure a fair comparison with~\cite{Saha2016}, 
we evaluate their offline tube generation method using the detection bounding boxes produced by the SSD net.
As in \cite{lin2014microsoft}, we report the mAP averaged over thresholds from 0.5 to 0.95 in steps of 0.05. 

\textbf{Improvement over the top performers}.
Results on {UCF101-24} are reported in Table~\ref{table:ucf101_results}.
%Table~\ref{table:ucf101_results} presents the results we obtained on UCF-101, and 
%compares them to the previous state-of-the-art~\cite{Saha2016,Weinzaepfel-2015,peng2016eccv,yu2015fast}.
In an online real-time setting we achieve an mAP of $70.2\%$ compared to $66.6\%$ reported by~\cite{Saha2016} 
at the standard IoU threshold of $\delta=0.2$.
{In non-real time mode, we observe a further performance improvement of around $3.3\%$, leading to a $73.5\%$ mAP, comparable
to the $73.5$ reported by the current top performer ~\cite{peng2016eccv}. 
The similar performance of our method (A$+$AF) to ~\cite{peng2016eccv} at $\delta=0.2$ suggests that 
SSD and the multi-region adaptation of Faster-RCNN by ~\cite{peng2016eccv} produce similar quality frame level detection boxes.}

\textbf{Performance under more realistic requirements}.
{Our method significantly outperforms ~\cite{Saha2016,peng2016eccv} at more meaningful higher detection thresholds $\delta=0.5$ or higher. 
For instance, we achieve a $46.2\%$ mAP at $\delta=0.5$ as opposed to the $32.1\%$ by ~\cite{peng2016eccv} and the $36.4\%$ by ~\cite{Saha2016}, 
an improvement of \textbf{14\%} and \textbf{9.8\%}, respectively.
This attests the superiority of our tube building algorithm when compared to those of ~\cite{peng2016eccv,Saha2016}.
In fact, our real-time mode (A $+$ RTF) performs better than both ~\cite{Saha2016,peng2016eccv} at $\delta=0.5$ or higher.}

% ----------------------------------------------------------------------------------------------------------------------------
% Fabio: this part seems important and should be expanded on/mentioned in the intro as well
% Cordelia is typically using a lot of hacking to get good results.
% Also, I do not underst what u means by applying detectors to whole bodies or single body parts, are they trained on single body parts? And where do they get the annotation for that?
% ----------------------------------------------------------------------------------------------------------------------------

It is important to note that, our proposed fusion method (\textit{union-set-fusion}) 
significantly outperforms \textit{boost-fusion} proposed by~\cite{Saha2016} on UCF101-24 dataset (see Table~\ref{table:ucf101_results}).
UCF-101 includes many co-occurring action instances, we can infer that the union-set fusion strategy 
improves the performance by providing a larger number of high confidence boxes from either the
appearance or the flow network.
When a single action is present in each video, as in 
JHMDB, \textit{boost-fusion} perform better (Table~\ref{table:jhmdb21_results}).
In the supplementary material we present a complete class-wise performance comparison of the two fusion strategies on both datasets.

%--------------------------------------------------------------------------------------------------------------------------
% UCF-101 results
%--------------------------------------------------------------------------------------------------------------------------
\begin{table}[t]
  %\vskip -3mm
  \centering
  \caption{S/T action localisation results (mAP) on untrimmed videos of UCF101-24 dataset in split1.}
  \vskip 1mm
  {\footnotesize
  \scalebox{0.97}{
  \begin{tabular}{lcccc}
  \toprule
  %\hline
  %IoU threshold $\delta$ 			 & 0.1   & 0.2    & 0.5 & 0.6 &0.5:0.95 \\ \midrule
  IoU threshold $\delta$ 			       & 0.2   & 0.5 & 0.75 &0.5:0.95 \\ \midrule
  Yu \etal~\cite{yu2015fast}${}^{\ddagger}$ 			           & 26.5  & --   & -- & --\\
  Weinzaepfel~\etal~\cite{Weinzaepfel-2015}${}^{\ddagger}$  & 46.8  & --   & -- & -- \\   
  Peng and Schmid~\cite{peng2016eccv}${}^{\dagger}$	 & \textbf{73.5}  & 32.1 & 02.7 & 07.3 \\
  Saha ~\etal~\cite{Saha2016}${}^{\dagger}$			           & 66.6  & 36.4 & 07.9 & 14.4 \\\midrule
  Ours-$\textnormal{Appearance (A)}^{*}$          & 69.8  & 40.9 & \textbf{15.5} & 18.7 \\
  Ours-$\textnormal{Real-time-flow (RTF)}^{*}$  	 & 42.5  & 13.9 & 00.5 & 03.3  \\ 
  Ours-$\textnormal{A $+$ RTF (boost-fusion)}^{*}$ & 69.7  & 41.9 & 14.1 & 18.4 \\
  Ours-$\textnormal{A $+$ RTF (union-set)}^{*}$ 	 & 70.2  & 43.0 & 14.5 & 19.2 \\\midrule
  Ours-$\textnormal{Accurate - flow (AF)}^{**}$   & 63.7  & 30.8 & 02.8 & 11.0 \\
  Ours-$\textnormal{A $+$ AF (boost-fusion)}^{**}$ & 73.0  & 44.0 & 14.1 & 19.2 \\
  Ours-$\textnormal{A $+$ AF (union-set)}^{**}$ 	 & \textbf{73.5}  & \textbf{46.3} & 15.0 & \textbf{20.4} \\\midrule
  SSD$+$~\cite{Saha2016} A $+$ AF (union-set)${}^{\dagger}$ & 71.7  & 43.3 & 13.2 & 18.6 \\
%   IoU threshold $\delta$ 			& 0.05  & 0.1   & 0.2    & 0.5 & 0.6\\ \midrule
%   Yu \etal~\cite{yu2015fast} 			& 49.90 & 42.80 & 26.50  & -- & --\\
%   Peng and Schmid~\cite{peng2016eccv} 		& 54.50 & 50.40 & 42.30  & -- &  -- \\
%   Weinzaepfel~\etal~\cite{Weinzaepfel-2015} 	& 54.28 & 51.68 & 46.77  & -- & --\\   
%   Saha ~\etal~\cite{Saha2016}			& 79.12 & 76.57 & 66.75  & 35.86 & 26.79 \\\midrule
  %\hline
  \bottomrule 
  %\vskip{1mm}
\multicolumn{5}{l}{${}^{\ddagger}$ These methods were using different annotations to~\cite{peng2016eccv,Saha2016} and ours \&} \\
\multicolumn{5}{l}{new annots available at \tiny{\url{https://github.com/gurkirt/corrected-UCF101-Annots}}} \\
\multicolumn{5}{l}{${}^{*}$ Incremental \& real-time \ \ ${}^{**}$ Incremental, non real-time \ \ ${}^{\dagger}$ Offline} % and non real-time}
  \end{tabular}
  }
  }
  %\vspace*{-\baselineskip}
  \label{table:ucf101_results} \vspace{-5mm}
\end{table} 

\textbf{Evaluation on J-HMDB-21}.
Table~\ref{table:jhmdb21_results} reports action detection results
averaged over the three splits of \textit{J-HMDB-21}, 
and compares them with those to our closest (offline) competitors.
\noindent Our framework outperforms the multi-stage approaches 
of ~\cite{Georgia-2015a,wangcvpr2016,Weinzaepfel-2015} in non real-time mode at the
standard IoU threshold of $0.5$, while it attains figures very close to those of
\cite{Saha2016,peng2016eccv} (73.8 versus 74.1 and 72.6, respectively) approaches, 
which make use of a two-stage Faster-RCNN.

Once again it is very important to point out that \cite{peng2016eccv} employs a battery of frame-level detectors, 
among which one based on strong priors on human body parts. 
Our approach does not make any prior assumption on the object(s)/actors(s) performing the action of interest, and is thus arguably more general-purpose.
%Still our approach is competitive with the best offline methods, while being online and real-time.

\subsection{Discussion}

\textbf{Contribution of the flow stream.} 
The optical flow stream is an essential part of the framework. % as it is needed to achieve high performance.
Fusing the real-time flow stream with the appearance stream (A$+$RTF mode) on UCF101-24 leads to a $2.1\%$ improvement at $\delta=0.5$. Accurate flow adds a further $3.3\%$.
A similar trend can be observed on JHMDB-21, where A$+$RTF gives a $5\%$ boost at $\delta=0.5$, 
and the A$+$RTF mode takes it further to $72\%$. 
It is clear from Table~\ref{table:ucf101_results} and Table~\ref{table:jhmdb21_results} 
that optical flow plays a much bigger role on the JHMDB dataset as compared to UCF101-24.
Real-time OF does not provide as big a boost as accurate flow, but still 
pushes the overall performance towards that of the top competitors, 
with the invaluable addition of real-time speed.}

{\textbf{Relative contribution of tube generation and SSD.} 
As anticipated
we evaluated the offline tube generation method of~\cite{Saha2016} 
using the detection bounding boxes produced by the SSD network,
to both provide a fair comparison and to understand each component's influence on performance.
The related results appear in the last row of Table~\ref{table:ucf101_results} and Table \ref{table:jhmdb21_results}.
\\
From comparing the figures in the last two rows of both tables
%It is clear from Table~\ref{table:ucf101_results} and Table ~\ref{table:jhmdb21_results}.
it is apparent that our online tube generation performs better than the
offline tube generation of \cite{Saha2016}, especially providing significant improvements at 
higher detection thresholds for both datasets.
We can infer that the increase in performance comes from both 
the higher-quality detections generated by SSD, 
as well as our new online tube generation method.
The fact that our tube genration is online, gready and outperforms offline methods, 
so it suggests that offline approaches has big room for improvements.

The reason for not observing a big boost due to the use of SSD on JHMDB may be 
its relatively smaller size, which does not allow us to leverage on the expressive power of SSD models.
Nevertheless, cross validating the CNNs' hyper-parameters (e.g. learning rate), 
might lead to further improvements there as well.

%It is noteworthy that our online tube linking able to outperform pervious state-of-the-art offline tube genration methods
%the SSD network might have not been trained well (overfit) on JHMDB dataset, which can explain the loss of performance as compared to Saha \etal~\cite{Saha2016}.
%\MS{make this explanation more convincing.}
%--- [SUMAN] - not using any more the union set fusion, so commenting the below ---
% Note that \textit{union-set} fusion does not perform as well as \textit{boost-fusion}, 
% probably because JHMDB videos contains only one action while \textit{union-set} 
% may produce two high-scoring tubes, yielding high scoring false positives.
%We show more extensive result on J-HMDB-21 dataset in supplementary material
%Sample detections on JHMDB are shown in Figure \ref{fig:J-HMDB_results_detection}.
\iffalse
\begin{figure}[t]
  \vskip -3mm
  \centering
  \includegraphics[width=0.98\textwidth]{figures/experiments/visual_JHMDB.pdf}
  \vskip -2mm
  \caption{\label{fig:J-HMDB_results_detection}
    {\small
      \textit{
        Sample space-time action localisation results on J-HMDB-21.
      }
    }
  }
  \vspace{-6mm}
\end{figure}
\fi

\begin{table}[t] %\vspace{-3mm}
\centering
\footnotesize
\caption{S/T Action localisation results (mAP) on J-HMDB-21.} 
%dataset, averaged over 3 splits.}
\vskip 2mm
%\begin{center}
\scalebox{0.97}{
\begin{tabular}{lcccc}
%\hline
\toprule
 IoU threshold $\delta$ 			                    & 0.2    & 0.5 & 0.75 & 0.5:0.95 \\ \midrule
%\hline\hline
%  Gkioxari and Malik~\cite{Georgia-2015a} 	& --    & 53.30  & --    & --  & -- \\
 Gkioxari and Malik~\cite{Georgia-2015a}${}^{\dagger}$ 	        & --   & 53.3  & --   & -- \\
 Wang~\etal~\cite{wangcvpr2016}${}^{\dagger}$   		            & --   & 56.4  & --   & -- \\
 Weinzaepfel~\etal~\cite{Weinzaepfel-2015}${}^{\dagger}$	      & 63.1 & 60.7  & --   & -- \\
 Saha ~\etal~\cite{Saha2016}${}^{\dagger}$ 			                & 72.6 & 71.5  & 43.3 & 40.0 \\
 Peng and Schmid~\cite{peng2016eccv}${}^{\dagger}$		          & \textbf{74.1} & \textbf{73.1}  & --   & -- \\\midrule
 Ours-$\textnormal{Appearance (A)}^{*}$                & 60.8 & 59.7  & 37.5 & 33.9 \\
 Ours-$\textnormal{Real-time-flow (RTF)}^{*}$  	      & 56.9 & 47.4  & 20.2 & 19.3 \\ 
 Ours-$\textnormal{A $+$ RTF (union-set)}^{*}$ 	        & 66.0 & 63.9  & 35.1 & 34.4 \\
 Ours-$\textnormal{A $+$ RTF (boost-fusion)}^{*}$	      & 67.5 & 65.0  & 36.7 & 38.8 \\\midrule
 Ours-$\textnormal{Accurate - flow (AF)}^{**}$  	      & 68.5 & 67.0  & 38.7 & 36.1 \\
 Ours-$\textnormal{A $+$ AF (union-set)}^{**}$ 	        & 70.8 & 70.1  & 43.7 & 39.7 \\
 Ours-$\textnormal{A $+$ AF (boost-fusion)}^{**}$	      & 73.8 & 72.0  & \textbf{44.5} & \textbf{41.6} \\\midrule
 SSD$+$~\cite{Saha2016} A $+$ AF (boost-fusion)${}^{\dagger}$   & 73.2 & 71.1  & 40.5 & 38.0 \\
\bottomrule 
\multicolumn{5}{l}{${}^{*}$ Incremental \& real-time \ \ ${}^{**}$ Incremental, non real-time \ \ ${}^{\dagger}$ Offline} % and non real-time}
\end{tabular}
}
%\vspace{-6mm}
%\vspace*{-\baselineskip}
\label{table:jhmdb21_results}
\vspace{-5mm}
\end{table} 

\subsection{Test time detection speed} %\vspace{-2mm}
\label{sec:detection-speed}
To support our claim to real time capability, 
we report the test time detection speed of our pipeline 
under all three types of input A (RGB), A$+$RTF (real-time flow), A $+$ AF (accurate flow) in
Table~\ref{table:detection_time_analysis}.
These figures
were generated using a desktop computer with an Intel Xeon CPU@2.80GHz (8 cores)
and two NVIDIA Titan X GPUs.
Real-time capabilities can be achieved by either not using optical flow (using only appearance (A) stream on one GPU)
or by computing real-time optical flow~\cite{kroeger2016fast} on a CPU in parallel
with two CNN forward passes on two GPUs.
For action tube generation (\S~\ref{subsec:action_tube}) we ran 8 CPU threads in parallel for each class.
We used the real-time optical flow algorithm ~\cite{kroeger2016fast} in a customised setting, 
with minimum number of pyramid levels set to $2$ instead of $3$, and 
patch overlap $0.6$ rather than $0.4$.
OF computation averages $\sim7$ ms per image. 

%To the best of our knowledge, we are the first to report test time detection speed for online spatio-temporal action localisation.
Table \ref{table:detection_time_analysis} also compares our detection speed to that reported by Saha \etal \cite{Saha2016}.
With an overall detection speed of 40 fps (when using RGB only) and 28 fps (when using also real time OF), 
our framework is able to detect multiple co-occurring action instances in real-time, while retaining very competitive performance.

\begin{table}[ht!]
\vskip -1mm
\centering
\caption{Test time detection speed.}% in frame per seconds (fps)}% on the  UCF101 datasets.}
\vskip 1mm
\resizebox{0.46\textwidth}{!}{
\begin{tabular}{l  c  c  c c}
\toprule
Framework modules & A & A$+$RTF & A$+$AF & \cite{Saha2016} \\\midrule
Flow computation (ms$^*$)  & -- & 7.0 & 110 & 110\\
Detection network time (ms$^*$)  & 21.8 & 21.8 & 21.8 & 145 \\
Tube generation time (ms$^*$)  & 2.5 & 3.0 & 3.0 & 10.0\\\midrule
Overall speed (fps$^{**}$ )  & 40 & 28 & 7 & 4 \\
%final speed with latency (fps$*$)
%\multirow{2}{*}{\parbox{4cm}{Detection speed with \\ single frame latency (fps$^*$ )}}  & \multirow{2}{*}{54} & \multirow{2}{*}{30} & \multirow{2}{*}{9} \\ \\
\bottomrule 
\multicolumn{3}{l}{$^*$ ms - milliseconds \ \ \ $^{**}$ fps - frame per second.} 
\end{tabular}
}
%\vspace*{-\baselineskip}
\vskip -4mm
\label{table:detection_time_analysis} %\vspace{5mm}
\end{table} 

\section{Conclusions and future plans} \label{sec:conclusions}
%% NO CHNAGES
We presented a novel online framework for action localisation and prediction able to 
%unlike existing state-of-the-art approaches which typically deal with single action classification and/or localisation problems on temporally trimmed videos,
address the challenges involved in concurrent multiple human action recognition, spatial localisation and temporal detection, in real time.
Thanks to an efficient deep learning strategy for the simultaneous detection and 
classification of region proposals and a new incremental action tube generation approach,
our method achieves superior performances compared to the previous state-of-the-art 
on early action prediction and online localisation, 
while outperforming the top offline competitors, in particular at high detection overlap.
%on the most challenging benchmark datasets,  and it is capable of handling multiple concurrent action instances and temporally untrimmed videos.
%Thanks to the incremental nature of the tube construction algorithm, our approach   can handle early action label prediction and localisation in an online fashion.
Its combination of high accuracy and fast detection speed at test time paves 
the way for its application to real-time applications such as autonomous driving, 
human robot interaction and surgical robotics, among others.

A number of future extensions can be envisaged. Motion vectors~\cite{zhangcvpr2016}, for instance, could be used in place of optical flow to achieve faster detection speeds. An even faster frame level detector, such as YOLO \cite{redmon2016yolo9000}, could be easily incorporated.
More sophisticated online tracking algorithms~\cite{wu2013online} for tube generation could be explored.
%In the medium term, however, we will pursue radically new deep network architectures able to regress entire tubes without having to rely on frame level detectors.

%\par\vfill\par
%\clearpage
\newpage
{\small
\bibliographystyle{ieee}
\bibliography{egbib}
}
\end{document}